\documentclass{article}

\PassOptionsToPackage{numbers, compress}{natbib}

\usepackage[preprint]{neurips_2021}

\usepackage[utf8]{inputenc} 
\usepackage[T1]{fontenc}    
\usepackage{hyperref}       
\usepackage{url}            
\usepackage{booktabs}       
\usepackage{nicefrac}       
\usepackage{microtype}      
\usepackage{xcolor}         
\usepackage[utf8]{inputenc}
\usepackage{setspace}

\usepackage{CJKutf8}
\usepackage{natbib}
\usepackage{mathtools}
\usepackage{mathrsfs}
\usepackage{graphicx}
\usepackage{amsmath}
\usepackage[ruled,vlined]{algorithm2e}

\title{Improving RNA Secondary Structure Design using Deep Reinforcement Learning}

\author{%
  Alexander Whatley$^*$ \qquad  Zhekun Luo$^*$ \qquad  Xiangru Tang\thanks{All authors contributed equally to this work.} \\
  Department of Electrical Engineering and Computer Sciences\\
  University of California, Berkeley\\

}

\begin{document}

\maketitle

\begin{abstract}
Rising costs in recent years of developing new drugs and treatments have led to extensive research in optimization techniques in biomolecular design. Currently, the most widely used approach in biomolecular design is directed evolution, which is a greedy hill-climbing algorithm that simulates biological evolution. In this paper, we propose a new benchmark of applying reinforcement learning to RNA sequence design, in which the objective function is defined to be the free energy in the sequence's secondary structure. In addition to experimenting with the vanilla implementations of each reinforcement learning algorithm from standard libraries, we analyze variants of each algorithm in which we modify the algorithm's reward function and tune the model's hyperparameters. We show results of the ablation analysis that we do for these algorithms, as well as graphs indicating the algorithm's performance across batches and its ability to search the possible space of RNA sequences. We find that our DQN algorithm performs by far the best in this setting, contrasting with, in which PPO performs the best among all tested algorithms.  Our results should be of interest to those in the biomolecular design community and should serve as a baseline for future experiments involving machine learning in molecule design.  
\end{abstract}

\section{Introduction}
The expanding scale and inherent complexity of biological data have encouraged a growing use of machine learning in biology to build informative and predictive models of the underlying biological processes \cite{andronescu2004new, gonzalez2015bayesian, alipanahi2015predicting, killoran2017generating, greener2021guide}.
Designing biological molecules more efficiently will have a great impact across many industries, from increasing the quality and yield of agricultural products, to lowering the price and development costs of new pharmaceutical drugs \cite{3b5860203cc143a8bbcd61da1e989c8f,biswas2018toward,angermueller2019model, biswas2021low}. However, the procedure for designing new molecules is a difficult optimization problem that relies on data from rounds of wet-lab experiments. Especially, the design task is more challenging for various classes of RNA structures as the size of the target structure is increased \cite{aguirre2007computational}. Currently, the most widely used approach in biomolecular design is directed evolution, which is a greedy hill-climbing algorithm that simulates biological evolution, in which the next experiment to be performed is based upon modifications of the highest performing molecules in the current experiment batch \cite{de2018utility}. Several machine learning-based approaches and benchmarks have been proposed that show improvements over directed evolution. These benchmarks compare results from algorithms based on reinforcement learning, generative models, Bayesian optimization, and other machine learning techniques, and then compare these algorithms' results to a baseline model based on a greedy algorithm, which simulates directed evolution.

In biomolecular design, we are given an objective function $f$ which measures some property of given input sequences $x$, and we wish to find the optimum $x$ that maximizes $f$ via an iterative algorithm \cite{matthies2012dynamics, andronescu2003rnasoft}. Adding more complexity, in an experimental setting, rounds of large batches of molecules will be made and evaluated, and the results will inform the next round of designed molecules. Any algorithm used to optimize the objective function $f$ will need to account for these design constraints \cite{lyngs1999fast,dowell2004evaluation,bindewald2011multistrand}. 

Currently, the industry standard for biomolecular design is directed evolution. Directed evolution seeks to mimic the biological process of evolution, in which a library of molecules are generated and screened, and then the top molecules are taken and modified in some way. These modified molecules then serve as the next batch of molecules to be tested and screened, and this process is repeated. 

Recent machine learning approaches to molecule design have shown advantages over greedy approaches such as directed evolution. Generative autoencoders and GANs have been used to generate SMILES strings of molecules with specific properties \cite{blaschke2018application} \cite{de2018molgan}. Deep Q-Learning was used to optimize SMILES strings for quantities such as logP and QED \cite{zhou2018optimization}. Most recently, in \cite{angermueller2019model}, the authors apply several state-of-the-art reinforcement learning algorithms to three benchmarks that they create. In these three benchmarks, sequences are evaluated based on the Ising model for protein structure, which can be computed without wet-lab experiments. 

We propose a new benchmark in the realm of RNA sequence design, in which sequences are optimized by the free energy in their secondary structure. This task is inspired by the important need of designing RNA primers for tasks such as CRISPR therapeutics \cite{chuai2018deepcrispr} which relies heavily on sequences with low free energy. Additionally, ViennaRNA \cite{hofacker2003vienna,lorenz2011viennarna} has been shown to accurately measure secondary structures in RNA sequences and calculate their free energy, allowing us to calculate our objective function.

\section{Methods}
We explore three different algorithms to solve the task of finding the RNA sequence with the highest fitness value. We first test the classic Deep Q-Network (DQN) approach for our task. Then to deal with the sparsity of the RNA sequences with high energy states, we also try Proximal Policy Optimization (PPO) to approximate the sparse reward landscape.  Besides, we also implemented a baseline model with a greedy search algorithm as a benchmark. We discuss all these three approaches in the following sub-sections. In order to train our agent, we also implemented a reinforcement learning environment for the RNA sequence. We then adapted the TensorFlow RL agents library to this RNA environment. 

\subsection{The Deep Q-Network (DQN) approach}
The classic Deep Q-network approach adapts the following update equation to learn to Q function, which implicitly represents a policy.

\begin{center}
$  Q(s,a) = (1-\alpha)*Q(s,a)~+~\alpha*(~R(s,a) + \gamma*\max_{a'} Q(s',a'))    $
\end{center}

We adapt the classic DQN algorithm to our RNA energy-optimization task. Our hypothesis is that the RNA structures with high energy states have some latent similarities in their structures. Thus we frame this problem as a reinforcement learning problem. We start from some initially random RNA sequences. Then we gradually change the RNA elements in sequences, one at a time. Each position of the RNA sequence can take four different values: G, U, A, and C.
 
We frame the "flip" in one position of the RNA sequence (from one to the other three) as an action step of our RL agent. Our environment returns the fitness value of the new sequence as the reward agent gets. 

Note that this setup is a bit different from the classic DQN setup. In a classic DQN problem, the agents aim to find a policy that generates the highest sum of rewards. In comparison, in our task, we aim to find the RNA sequences with the highest energy state. In other words, our objective is to learn the policy which generates the highest "max" rewards along its trajectory. However, our results show that the DQN algorithm works quite well for our tasks. Intuitively, the RNA sequences encode its biological structure, and the relationship between sequences and fitness value is not random. The RNA sequence with high energy states can be viewed as generated by specific underlying code associated with high fitness. These sequences share some similarities in their latent features. Thus we can expect that when we arrive at the RNA sequences with high rewards, the sum of rewards along its trajectory will also have a high probability to be high than average. Our experiment results confirm our hypothesis regarding the sequence-generating process.

We used the classic DQN algorithm with a replay buffer, along with prioritized experience replay \cite{schaul2015prioritized}. We do this as the top sequences become more and more sparse At the beginning of each iteration, we make some moves in the environment according to the current policy and collect the trajectories. We add these trajectories to our buffer and train for several iterations from there. We keep a dictionary to document the seen sequences in the current trajectory. We punish the agents by directly terminating the trajectory whenever it steps into the same environment twice. This way, we prevent the agents from going into a loop.

Also, note that our environment state is specified via its initial(random) state. We reset the RNA environment to different initial states at the beginning of each iteration. Our algorithm is detailed below.

\begin{algorithm}[H]
\SetAlgoLined
\caption{Deep Q-Network Approach}
 Replay buffer \( \mathcal{B} \)\\ 
  Seen sequence set \( \mathcal{S} \)\\ 
 Fitness function $f$ \\
   \vskip 0.05in
\For{E Epoches} {
  \vskip 0.1in
 \textbf{Collect Data}:\\  
 
 \For{i in Range(M)} {
    \# take some action in the environment  \\
    Take flip $A_i$ on sequence $S_i$ according to the current policy \\
    Receive the new sequence $S'_i$ \\ 
    \# if we step into a loop  \\
    \If {$S'_i$ in $\mathcal{S}$}
    {
    \# clear the seen set  \\
    \( \mathcal{S} \) = \O \\
    Return $\mathcal{T}$
    }  
    \Else{
    $\#$ calculate the fitness  \\
    $R_i$ = $f(S'_i)$ \\
    $\#$ add too seen set \\
    
    $S'_i$ $\rightarrow$ $\mathcal{S}$ \\
    $\#$ add too replay buffer \\
    ($S_i, A_i, S'_i, R_i$) $\rightarrow$ $\mathcal{B}$ \\
    }
    
}
  \vskip 0.1in
 
 \textbf{Training}:\\  
 
 \For{j in Range(N)} {
    $\#$ sample trajectory  \\
    ($S_i, A_i, S'_i, R_i$) $\leftarrow$ $(\mathcal{B})$\\
    $\#$ fit network to target via SGD \\
   $\theta$ $\leftarrow \theta + \alpha* \nabla \theta$ \\
    
}}
\end{algorithm}

\subsection{The Proximal Policy Optimization(PPO) approach}

We also explored our task with another popular deep RL method: Proximal Policy Optimization (PPO). Intuitively, The PPO algorithm is doing importance sampling. The PPO algorithm regularizes the gradient by controlling the distribution's divergence as a constraint. Usually, this is framed as a Lagrange multiplier. This way of approximating the direction of the gradient usually ensures the estimated gradient is closer to the true gradient as the constraints are enforced.  Specially, it optimizes the following objective:

\begin{equation}
\centering
 \theta' \leftarrow \underset{\theta'}{\operatorname{argmax}} \nabla_{\theta} \bar{A}(\theta)^T(\theta' - \theta)  
\end{equation}
\vskip 0.1in
But the above policy approximation is accurate only is the distribution shift of the policy is not big, concretely the constraint can be written as:

\vskip 0.1in
\begin{equation}
\centering
  D_{KL}(\pi_{\theta'}(a_{t}\mid s_{t}) . \mid\mid \pi_{\theta}(a_{t}\mid s_{t})  ) <= \epsilon 
\end{equation}
\vskip 0.1in

Here the distribution shift of the policy is bounded by a hyper-parameter $\epsilon $ In practice, sampling methods are used to calculate the gradient. 

Note that the above equations are from the lecture slides of the Berkeley cs285 Deep Reinforcement Learning course lecture slides. For detailed information regarding the Proximal Policy Optimization algorithm, we refer the readers to a series of papers discussing these types of algorithms in detail. Our PPO agent is implemented based on the TensorFlow RL agent library. We modified its code to adapt to our problem setting.

The detailed algorithm is described below. It used a similar training process as in the Deep-Q Network algorithm. One difference is that here in the data collection process, we add a parameter called $Try\_Iteration$. When we sample the trajectories to put into the replay buffer as training data, we need to prevent the agents from going into the same state repeatedly. While some papers introduced a punishment reward as a regularizer, in the Deep-Q Network algorithm, we use direct termination as a tool to prevent the loop. It works well in the DQN setting. However, the same methods seem to fail in the PPO setting. We observed the performance is quite bad, with a plain terminating in the PPO experiments. So instead, we explore another "try again" policy.

\begin{algorithm}[htp]
\SetAlgoLined
\caption{Proximal Policy Optimization Approach}

 Replay buffer \( \mathcal{B} \)\\ 
  Seen sequence set \( \mathcal{S} \)\\ 
 Fitness function $f$ \\
   \vskip 0.05in
\For{E Epoches} {
  \vskip 0.1in
 \textbf{Collect Data}:\\  
 
 \For{i in Range(M)} {
    $try\_iter \leftarrow 0$ \\
    Take flip $A_i$ on sequence $S_i$ according to the current policy \\
    Receive the new sequence $S'_i$ \\ 
    
    \While{Try\_Iter $\leq$	 Max\_Iter and $S'_i$ in \( \mathcal{S} \)}  
    {Take flip $A_i$ on sequence $S_i$ according to the current policy \\
    Receive the new sequence $S'_i$ \\ 
    Try\_Iter += 1\\
     \If {$Try\_Iter == Max\_iter$} 
    {\( \mathcal{S} \) = $\O$ \\
    Return \( \mathcal{T} \)}
    
    }
    
    
    
    \# calculate the fitness  \\
    $R_i$ = $f(S'_i)$ \\
    \# add too seen set \\
    $S'_i$ $\rightarrow$ \( \mathcal{S} \) \\
    \# add too replay buffer \\
    $(S_i, A_i, S'_i, R_i)$ $\rightarrow$ \( \mathcal{B} \) \\

}
  \vskip 0.1in
 
 \textbf{Training}:\\  
 
 \For{j in Range(N)} {
    \# sample trajectory  \\
    ($S_i, A_i, S'_i, R_i$) $\leftarrow$ \( \mathcal{B} \) \\
    \# fit network to target via SGD \\
    $theta$ $\leftarrow$ $theta$ + $\alpha*$ $\nabla$ $\theta $
    
}
 }
\end{algorithm}

Concretely, if the agents have seen a state in the same trajectory before, instead of terminating directly, the agents would try to find another action that leads to an unexplored state. It will keep trying until it steps into an unseen state or it has tried more than $Max\_Iter$ times but still fails to find a valid move.

We modified the environment setting because empirically, we find the PPO algorithm is sensitive to terminating. It will converge to local optima very fast if the direct-termination regularizer is applied. With this modification, are agent achieves 2 times performance as the original environment setting. In the experiment result section, we conduct the ablation study on different methods to punish the looping behavior and discuss their corresponding influence in terms of converging patterns and average performance. 

\subsection{Baseline: the greedy search approach}
In order to compare our results, we also implemented a baseline algorithm and tested its performance in our task. It is a simple hill-climbing greedy search algorithm. We benchmark our PPO and DQN results based on this model. We start by generating some random RNA sequences and putting them into a buffer \( \mathcal{B} \) . For each iteration, we sample a batch of sequences from the buffer and iteratively perform some random mutations on the sequences. Then we calculate the fitness value of the modified RNA  structures according to our metrics. We only maintain the generated sequences with the top fitness scores and put them back into the buffer. The intermediate mutation results are discarded. This algorithm stops when we cannot make further improvements on the fitness values (local optimum), or when we reach the maximum number of iterations. The algorithm is detailed below. We include the result of the greedy baseline model in the experiment result section.

\begin{algorithm}[htp]
 \caption{Baseline Greedy Search Model}
\SetAlgoLined

 Mutation function \( \mathcal{M} \)\\ 
 Fitness function $f$ \\
 Sample $X_i$ with fitness score $y_i$ \\
 \vskip 0.1in
 \textbf{Initialization}:\\  
 \# random sample $N$ sequence and put it in buffer \( \mathcal{B} \) \\ 
 \( \mathcal{B} \)  $\leftarrow$ $X_i$ for $i = 1~...~N$ \\
 \vskip 0.1in
 \While{ iteration $m <= maximum~M$ }{
    \# random sample sequences from buffer \\
    $X_i$ $\leftarrow$ \( \mathcal{B} \) for $i = 1~...~n$ \\
    \vskip 0.05in
    \For{i in Range(n)} {
    \# do random mutation  \\
    $X_i'$ = \( \mathcal{M} \) $(X_i)$ for $i = 1~...~n$ \\
    \# calculate fitness score via fitness function\\
    $y_i$ = $f(X_i)$ \\
    $y_i'$ = $f(X_i')$ \\
    \# take the max one \\
    $X_i$ = $X_i'$ \textbf{if} $y_i'$ $>$ $y_i'$ \textbf{else} $X_i$
    }
    \vskip 0.05in
    \# put back to buffer \\
    \( \mathcal{B} \)  $\leftarrow$ $X_i$ for $i = 1~...~n$
 }
\end{algorithm}

\section{Experiment and Analysis}

In this section, we describe the results for the above models. In 3.1, we first reported the result for the plain greedy model, which acts as a baseline for our task. It achieves decent performance, and we attribute it to the inner pattern of the biology structure.  In 3.2-3.3, We report the results for both the Proximal Policy Optimization(PPO) and the Deep Q-Network(DQN) approach. We benchmark them against the baseline. We analyze their learning behavior and performance. We further discuss some potential improvements. Note that currently, the PPO model is not working quite well. We put forward some hypotheses and discuss them in detail in the following section 3.3. In 3.4, we conduct an ablation study in terms of the method we embedded into the environment setup to prevent looping. We confirm some intuition we discuss early in the methods section intuitively. We compare the result for different regularizing methods in terms of the converging pattern and also average performance. We conclude that the "try again" regularizer works best.

\subsection{Baseline result}

We used the greedy search algorithm as a baseline for our task. The agents iteratively try to find the next action the maximizes the fitness score, which is framed as a reward in an RL setting. The result we get for this plain greedy model is:

\begin{figure}[h]
\centering
\includegraphics[scale=0.35]{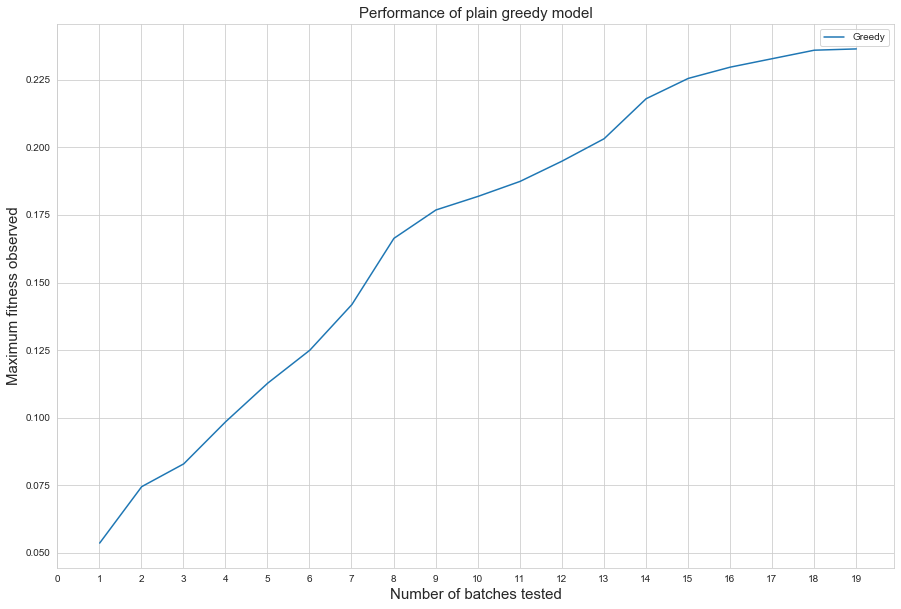}
\caption{The performance of the greedy model}
\end{figure}

We see that the baseline greedy algorithm performs quite well. It implies that the inner biological formation of RNA follows some pattern. And an agent with the plain greedy algorithm will go into a local optimum with a decent fitness score.

\subsection{Deep-Q Network approach}

One directly uses the fitness score as the reward 
(namely DQR PER) , another one modifies the reward function to punish the agents if it sees the same state for many times(namely DQR PER reward freq). The reward for the latter one is designed according to the following formula: \\

\vskip 0.1in
\begin{center}

$  R(s,a)$  = \( \mathcal{F} \)$(s')$ + $\alpha$*\( \mathcal{N} \)$(s') $  

\end{center}
\vskip 0.1in

Where \( \mathcal{F} \) is the function to calculate the fitness score in the new state $s'$, and \( \mathcal{N} \) counts how many times the agent has seen the state $s'$ before, in order to punish the agents from going into a loop. The $\alpha$ is a hyper-parameter to control the loss ratio and is set to be 0.1 through grid search. 

The learning curve in terms of optimal results for these two agents during training are:  

\begin{figure}[h]
\centering
\includegraphics[scale=0.35]{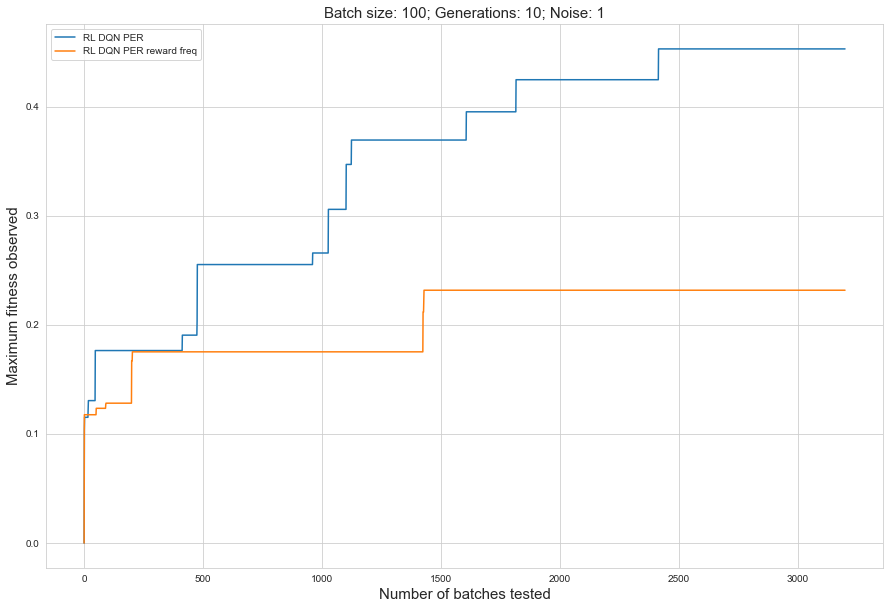}
\caption{The best result learning curve}
\end{figure}

While the learning curve in terms of average results for these two agents during training are the following. The performance fluctuates a bit, but is still relatively stable.

\begin{figure}[h]
\centering
\includegraphics[scale=0.35]{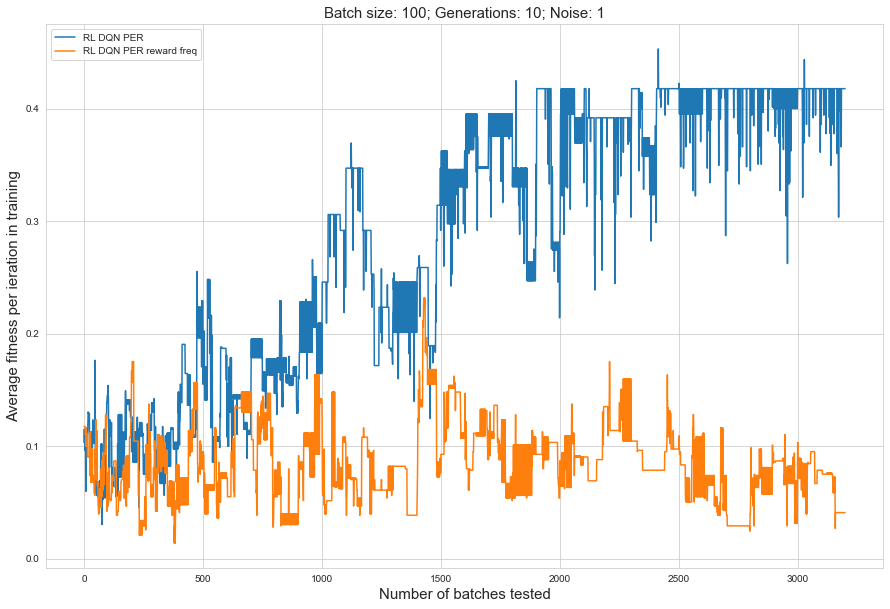}
\caption{The average result learning curve}
\end{figure}

We see that the agents with the plain reward function outperform the agent with the regularization to punish the looping behavior by almost two times in terms of the optimal result achieved. Thus we conclude that using a punishment factor in the reward function is not an effective way to prevent looping. That is probably due to the setting of reinforcement learning: the agents regard the state as bad we modify the reward function. But in fact, the state itself is not bad. Instead, the looping behavior is bad. 

Specifically, we put the trajectory into the replay buffer. Then these trajectory is decomposed into independent($S_i, A_i, S'_i, R_i$) pairs. When training the agent via replay buffer, we sample single ($S_i, A_i, S'_i, R_i$) samples. The agents do not see the full trajectory. It did not know about the past of the state. It would not know this action is bad because it has seen it before. Instead, it would conclude that the action itself is bad.  This does not align with our motivation to introduce the punishment factor.

\subsection{Proximal Policy Optimization approach}

We tested the Proximal Policy Optimization Approach(PPO) methods in our agents. We find that PPO does not work well in our task. The performance is lower than the baseline model. Also, it introduces a high variance. In order to demonstrate the high variance it creates, we used scatter plots here to exhibit the agent's behavior. 

\begin{figure}[h]
\centering
\includegraphics[scale=0.35]{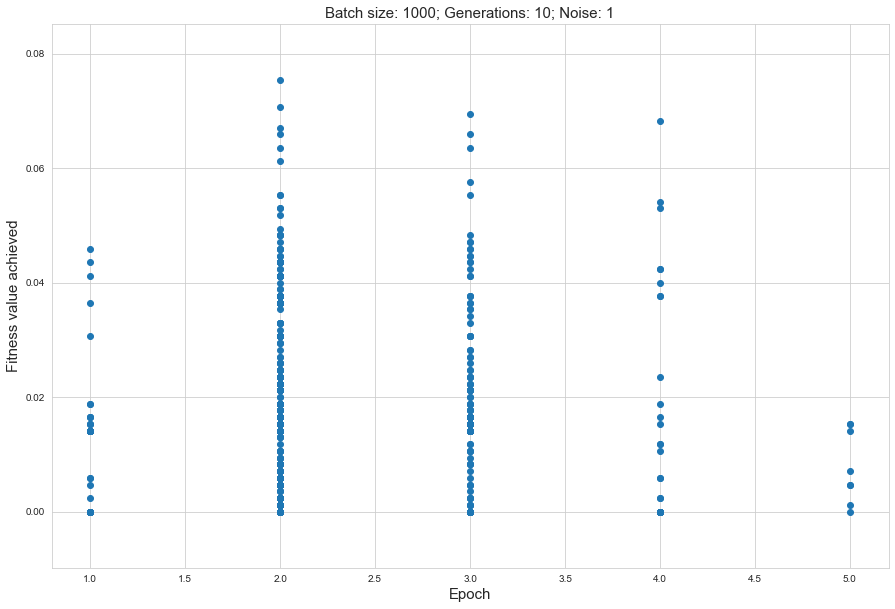}
\caption{The scatter plot for the Proximal Policy Optimization approach}
\end{figure}

We have several hypotheses of why PPO fails in our task. One is that PPO is designed to solve the policy distribution misteach problem when calculating the gradient because we are also updating our policy in the training process at the same time. However, such a mismatch problem might not be crucial in our problem setting because we have a relatively simple task. The search space for the policy is small. In other words, it takes a small space to encode a policy. This is empirically confirmed as we find that using a simple shadow neural network works best. Intuitively, since the policy search space is small, the distribution mismatch when updating policy is also small. Thus the advantage of the PPO algorithm is not utilized in this problem.

On the other hand, the PPO approach heavily relies on sampling methods to approximate the gradient direction. This will introduce high variance, which became fatal in our problem setting. Because despite a simple policy space, the action space is huge -- the agents can take action to flip any RNA position. Thus the probability of most of the action in our problem is close to zero. Therefore if we introduce high variance into training, in the early unstable stage of training, the agent will receive noisy signals. Specifically, if one action gets incorrectly boosted, it will soon outperform all other actions(since their probability is all close to zero). And such mistakes will get reinforced in the later training process.

We also suspect that the regularization method we use in DQN does not apply here. Also, the above result shows that the agents with rewards modified to prevent the looping behavior perform drastically worse. Empirically it seems that the PPO method is more sensitive to the anti-looping regularization method we use. Thus we do ablation studies on different regularization approaches and test them in the following section.

\subsection{Ablation study for the PPO regularizer}

In addition to the previously mentioned "terminating" approach, in which the environment directly terminates a trajectory if the agent sees the same states in the same trajectory twice, here we explore another approach to prevent looping. When the agent sees a seen state, instead of terminating, it will keep trying until it either steps into an unseen state or exceeds the maximum iteration limits. We compare the result in these two approaches in the following plot:

\begin{figure}[h]
\centering
\includegraphics[scale=0.35]{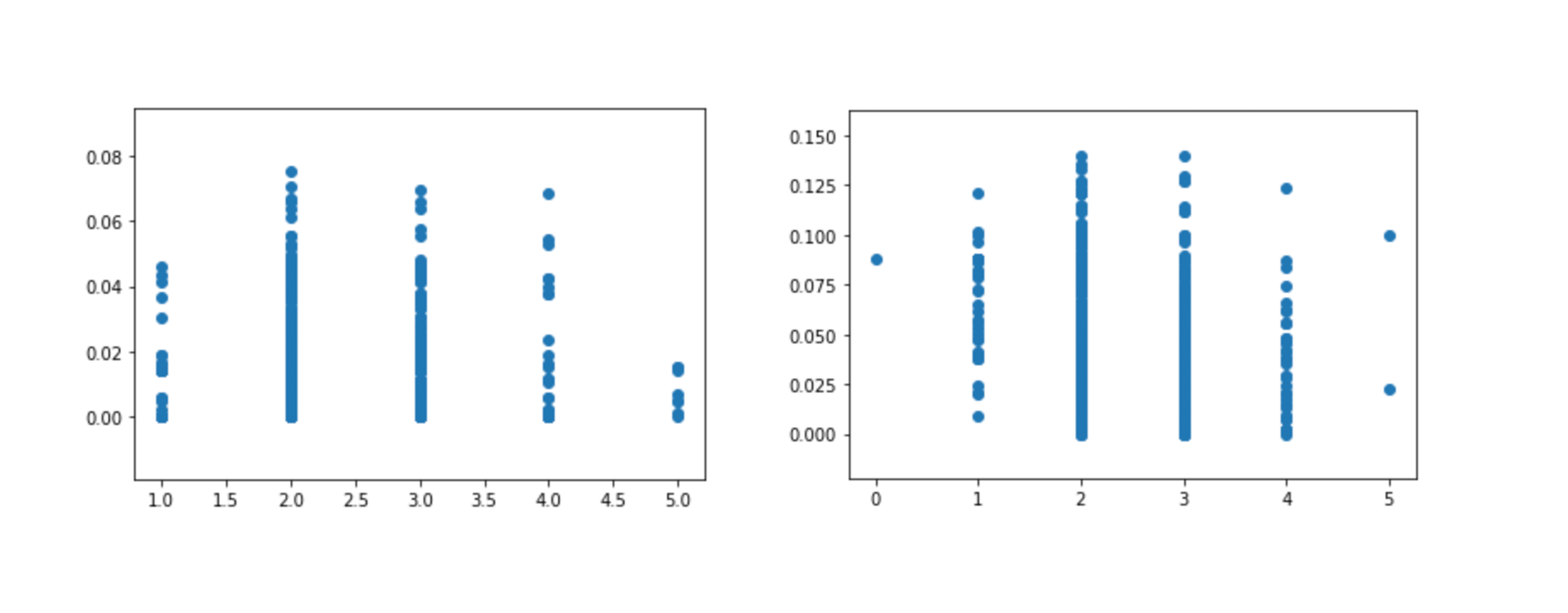}
\caption{Terminating approach (left) and "Try again" approach (right) }
\end{figure}

The agents with the "Try again" approach performs better. Our reasoning is similar to the one for failure using reward modification in section 3.2. Because if we terminate the state, the environment is explicitly telling the agents that the state is bad. But in fact, only looping is bad. Using the "Try again " policy better models our motivation.

\section{Conclusion and Future work}

In this paper, we propose a new benchmark of applying reinforcement learning to RNA sequence design, in which the objective function is defined to be the free energy in the sequence's secondary structure. This is motivated by the availability of the ViennaRNA library, which is able to measure RNA secondary structure and free energy, allowing us to obtain the reward. Additionally, the search space of possible RNA sequences, while still exponential in size, is exponentially smaller than those of SMILES strings or protein sequences, thus making this benchmark easier to evaluate. Designing an RNA sequence with an optimal secondary structure is important in many bioengineering tasks, such as in designing CRISPR primers \cite{yan2015crispr, reuter2010rnastructure, anderson2016principles}.

We test state-of-the-art deep reinforcement learning algorithms, namely Proximal Policy Optimization (PPO) and Deep Q-Networks (DQN), on RNA design tasks. For a baseline comparison, we show results for a basic greedy hill-climbing algorithm, which simulates the directed evolution process. In addition to experimenting with the vanilla implementations of each reinforcement learning algorithm from standard libraries, we analyze variants of each algorithm in which we modify the algorithm's reward function and tune the model's hyperparameters. We show results of the ablation analysis that we do for these algorithms, as well as graphs indicating the algorithm's performance across batches and its ability to search the possible space of RNA sequences. We find that our DQN algorithm performs by far the best in this setting, contrasting with \cite{angermueller2019model}, in which PPO performs the best among all tested algorithms.

Our results should be of interest to those in the biomolecular design community and should serve as a baseline for future experiments involving machine learning in molecule design.  We have shown in our experiments that reinforcement learning is a powerful alternative to greedy hill climbing or directed evolution in RNA design. In our setting, DQN performs significantly better than greedy does and thus is of significant interest to those in the biomolecular design community. This result is interesting as it contrasts with \cite{angermueller2019model}, in which PPO performs the best among all algorithms. To further understand the power of the application of reinforcement learning in biomolecular design, we will need to experiment on other biological design tasks and explore a larger repertoire of algorithms.

\bibliographystyle{achemso}
\bibliography{re.bib}

\providecommand{\latin}[1]{#1}
\makeatletter
\providecommand{\doi}
  {\begingroup\let\do\@makeother\dospecials
  \catcode`\{=1 \catcode`\}=2 \doi@aux}
\providecommand{\doi@aux}[1]{\endgroup\texttt{#1}}
\makeatother
\providecommand*\mcitethebibliography{\thebibliography}
\csname @ifundefined\endcsname{endmcitethebibliography}
  {\let\endmcitethebibliography\endthebibliography}{}
\begin{mcitethebibliography}{26}
\providecommand*\natexlab[1]{#1}
\providecommand*\mciteSetBstSublistMode[1]{}
\providecommand*\mciteSetBstMaxWidthForm[2]{}
\providecommand*\mciteBstWouldAddEndPuncttrue
  {\def\EndOfBibitem{\unskip.}}
\providecommand*\mciteBstWouldAddEndPunctfalse
  {\let\EndOfBibitem\relax}
\providecommand*\mciteSetBstMidEndSepPunct[3]{}
\providecommand*\mciteSetBstSublistLabelBeginEnd[3]{}
\providecommand*\EndOfBibitem{}
\mciteSetBstSublistMode{f}
\mciteSetBstMaxWidthForm{subitem}{(\alph{mcitesubitemcount})}
\mciteSetBstSublistLabelBeginEnd
  {\mcitemaxwidthsubitemform\space}
  {\relax}
  {\relax}

\bibitem[Andronescu \latin{et~al.}(2004)Andronescu, Fejes, Hutter, Hoos, and
  Condon]{andronescu2004new}
Andronescu,~M.; Fejes,~A.~P.; Hutter,~F.; Hoos,~H.~H.; Condon,~A. A new
  algorithm for RNA secondary structure design. \emph{Journal of molecular
  biology} \textbf{2004}, \emph{336}, 607--624\relax
\mciteBstWouldAddEndPuncttrue
\mciteSetBstMidEndSepPunct{\mcitedefaultmidpunct}
{\mcitedefaultendpunct}{\mcitedefaultseppunct}\relax
\EndOfBibitem
\bibitem[Gonzalez \latin{et~al.}(2015)Gonzalez, Longworth, James, and
  Lawrence]{gonzalez2015bayesian}
Gonzalez,~J.; Longworth,~J.; James,~D.~C.; Lawrence,~N.~D. Bayesian
  optimization for synthetic gene design. \emph{arXiv preprint
  arXiv:1505.01627} \textbf{2015}, \relax
\mciteBstWouldAddEndPunctfalse
\mciteSetBstMidEndSepPunct{\mcitedefaultmidpunct}
{}{\mcitedefaultseppunct}\relax
\EndOfBibitem
\bibitem[Alipanahi \latin{et~al.}(2015)Alipanahi, Delong, Weirauch, and
  Frey]{alipanahi2015predicting}
Alipanahi,~B.; Delong,~A.; Weirauch,~M.~T.; Frey,~B.~J. Predicting the sequence
  specificities of DNA-and RNA-binding proteins by deep learning. \emph{Nature
  biotechnology} \textbf{2015}, \emph{33}, 831--838\relax
\mciteBstWouldAddEndPuncttrue
\mciteSetBstMidEndSepPunct{\mcitedefaultmidpunct}
{\mcitedefaultendpunct}{\mcitedefaultseppunct}\relax
\EndOfBibitem
\bibitem[Killoran \latin{et~al.}(2017)Killoran, Lee, Delong, Duvenaud, and
  Frey]{killoran2017generating}
Killoran,~N.; Lee,~L.~J.; Delong,~A.; Duvenaud,~D.; Frey,~B.~J. Generating and
  designing DNA with deep generative models. \emph{arXiv preprint
  arXiv:1712.06148} \textbf{2017}, \relax
\mciteBstWouldAddEndPunctfalse
\mciteSetBstMidEndSepPunct{\mcitedefaultmidpunct}
{}{\mcitedefaultseppunct}\relax
\EndOfBibitem
\bibitem[Greener \latin{et~al.}(2021)Greener, Kandathil, Moffat, and
  Jones]{greener2021guide}
Greener,~J.~G.; Kandathil,~S.~M.; Moffat,~L.; Jones,~D.~T. A guide to machine
  learning for biologists. \emph{Nature Reviews Molecular Cell Biology}
  \textbf{2021}, 1--16\relax
\mciteBstWouldAddEndPuncttrue
\mciteSetBstMidEndSepPunct{\mcitedefaultmidpunct}
{\mcitedefaultendpunct}{\mcitedefaultseppunct}\relax
\EndOfBibitem
\bibitem[Chaudhury \latin{et~al.}(2010)Chaudhury, Lyskov, and
  Gray]{3b5860203cc143a8bbcd61da1e989c8f}
Chaudhury,~S.; Lyskov,~S.; Gray,~J. PyRosetta: A script-based interface for
  implementing molecular modeling algorithms using Rosetta.
  \emph{Bioinformatics} \textbf{2010}, \emph{26}, 689--691, Funding
  Information: Funding: PyRosetta was funded through National Institute of
  Health (R01-GM73151 and R01-GM078221); National Science Foundation CAREER
  Grant CBET (0846324).\relax
\mciteBstWouldAddEndPunctfalse
\mciteSetBstMidEndSepPunct{\mcitedefaultmidpunct}
{}{\mcitedefaultseppunct}\relax
\EndOfBibitem
\bibitem[Biswas \latin{et~al.}(2018)Biswas, Kuznetsov, Ogden, Conway, Adams,
  and Church]{biswas2018toward}
Biswas,~S.; Kuznetsov,~G.; Ogden,~P.~J.; Conway,~N.~J.; Adams,~R.~P.;
  Church,~G.~M. Toward machine-guided design of proteins. \emph{bioRxiv}
  \textbf{2018}, 337154\relax
\mciteBstWouldAddEndPuncttrue
\mciteSetBstMidEndSepPunct{\mcitedefaultmidpunct}
{\mcitedefaultendpunct}{\mcitedefaultseppunct}\relax
\EndOfBibitem
\bibitem[Angermueller \latin{et~al.}(2019)Angermueller, Dohan, Belanger,
  Deshpande, Murphy, and Colwell]{angermueller2019model}
Angermueller,~C.; Dohan,~D.; Belanger,~D.; Deshpande,~R.; Murphy,~K.;
  Colwell,~L. Model-based reinforcement learning for biological sequence
  design. International conference on learning representations. 2019\relax
\mciteBstWouldAddEndPuncttrue
\mciteSetBstMidEndSepPunct{\mcitedefaultmidpunct}
{\mcitedefaultendpunct}{\mcitedefaultseppunct}\relax
\EndOfBibitem
\bibitem[Biswas \latin{et~al.}(2021)Biswas, Khimulya, Alley, Esvelt, and
  Church]{biswas2021low}
Biswas,~S.; Khimulya,~G.; Alley,~E.~C.; Esvelt,~K.~M.; Church,~G.~M. Low-N
  protein engineering with data-efficient deep learning. \emph{Nature Methods}
  \textbf{2021}, \emph{18}, 389--396\relax
\mciteBstWouldAddEndPuncttrue
\mciteSetBstMidEndSepPunct{\mcitedefaultmidpunct}
{\mcitedefaultendpunct}{\mcitedefaultseppunct}\relax
\EndOfBibitem
\bibitem[Aguirre-Hern{\'a}ndez \latin{et~al.}(2007)Aguirre-Hern{\'a}ndez, Hoos,
  and Condon]{aguirre2007computational}
Aguirre-Hern{\'a}ndez,~R.; Hoos,~H.~H.; Condon,~A. Computational RNA secondary
  structure design: empirical complexity and improved methods. \emph{BMC
  bioinformatics} \textbf{2007}, \emph{8}, 1--16\relax
\mciteBstWouldAddEndPuncttrue
\mciteSetBstMidEndSepPunct{\mcitedefaultmidpunct}
{\mcitedefaultendpunct}{\mcitedefaultseppunct}\relax
\EndOfBibitem
\bibitem[de~Visser \latin{et~al.}(2018)de~Visser, Elena, Fragata, and
  Matuszewski]{de2018utility}
de~Visser,~J. A.~G.; Elena,~S.~F.; Fragata,~I.; Matuszewski,~S. The utility of
  fitness landscapes and big data for predicting evolution. 2018\relax
\mciteBstWouldAddEndPuncttrue
\mciteSetBstMidEndSepPunct{\mcitedefaultmidpunct}
{\mcitedefaultendpunct}{\mcitedefaultseppunct}\relax
\EndOfBibitem
\bibitem[Matthies \latin{et~al.}(2012)Matthies, Bienert, and
  Torda]{matthies2012dynamics}
Matthies,~M.~C.; Bienert,~S.; Torda,~A.~E. Dynamics in sequence space for RNA
  secondary structure design. \emph{Journal of chemical theory and computation}
  \textbf{2012}, \emph{8}, 3663--3670\relax
\mciteBstWouldAddEndPuncttrue
\mciteSetBstMidEndSepPunct{\mcitedefaultmidpunct}
{\mcitedefaultendpunct}{\mcitedefaultseppunct}\relax
\EndOfBibitem
\bibitem[Andronescu \latin{et~al.}(2003)Andronescu, Aguirre-Hernandez, Condon,
  and Hoos]{andronescu2003rnasoft}
Andronescu,~M.; Aguirre-Hernandez,~R.; Condon,~A.; Hoos,~H.~H. RNAsoft: a suite
  of RNA secondary structure prediction and design software tools.
  \emph{Nucleic acids research} \textbf{2003}, \emph{31}, 3416--3422\relax
\mciteBstWouldAddEndPuncttrue
\mciteSetBstMidEndSepPunct{\mcitedefaultmidpunct}
{\mcitedefaultendpunct}{\mcitedefaultseppunct}\relax
\EndOfBibitem
\bibitem[Lyngs \latin{et~al.}(1999)Lyngs, Zuker, and Pedersen]{lyngs1999fast}
Lyngs,~R.~B.; Zuker,~M.; Pedersen,~C. Fast evaluation of internal loops in RNA
  secondary structure prediction. \emph{Bioinformatics (Oxford, England)}
  \textbf{1999}, \emph{15}, 440--445\relax
\mciteBstWouldAddEndPuncttrue
\mciteSetBstMidEndSepPunct{\mcitedefaultmidpunct}
{\mcitedefaultendpunct}{\mcitedefaultseppunct}\relax
\EndOfBibitem
\bibitem[Dowell and Eddy(2004)Dowell, and Eddy]{dowell2004evaluation}
Dowell,~R.~D.; Eddy,~S.~R. Evaluation of several lightweight stochastic
  context-free grammars for RNA secondary structure prediction. \emph{BMC
  bioinformatics} \textbf{2004}, \emph{5}, 1--14\relax
\mciteBstWouldAddEndPuncttrue
\mciteSetBstMidEndSepPunct{\mcitedefaultmidpunct}
{\mcitedefaultendpunct}{\mcitedefaultseppunct}\relax
\EndOfBibitem
\bibitem[Bindewald \latin{et~al.}(2011)Bindewald, Afonin, Jaeger, and
  Shapiro]{bindewald2011multistrand}
Bindewald,~E.; Afonin,~K.; Jaeger,~L.; Shapiro,~B.~A. Multistrand RNA secondary
  structure prediction and nanostructure design including pseudoknots.
  \emph{ACS nano} \textbf{2011}, \emph{5}, 9542--9551\relax
\mciteBstWouldAddEndPuncttrue
\mciteSetBstMidEndSepPunct{\mcitedefaultmidpunct}
{\mcitedefaultendpunct}{\mcitedefaultseppunct}\relax
\EndOfBibitem
\bibitem[Blaschke \latin{et~al.}(2018)Blaschke, Olivecrona, Engkvist, Bajorath,
  and Chen]{blaschke2018application}
Blaschke,~T.; Olivecrona,~M.; Engkvist,~O.; Bajorath,~J.; Chen,~H. Application
  of generative autoencoder in de novo molecular design. \emph{Molecular
  informatics} \textbf{2018}, \emph{37}, 1700123\relax
\mciteBstWouldAddEndPuncttrue
\mciteSetBstMidEndSepPunct{\mcitedefaultmidpunct}
{\mcitedefaultendpunct}{\mcitedefaultseppunct}\relax
\EndOfBibitem
\bibitem[De~Cao and Kipf(2018)De~Cao, and Kipf]{de2018molgan}
De~Cao,~N.; Kipf,~T. MolGAN: An implicit generative model for small molecular
  graphs. \emph{arXiv preprint arXiv:1805.11973} \textbf{2018}, \relax
\mciteBstWouldAddEndPunctfalse
\mciteSetBstMidEndSepPunct{\mcitedefaultmidpunct}
{}{\mcitedefaultseppunct}\relax
\EndOfBibitem
\bibitem[Zhou \latin{et~al.}(2018)Zhou, Kearnes, Li, Zare, and
  Riley]{zhou2018optimization}
Zhou,~Z.; Kearnes,~S.; Li,~L.; Zare,~R.~N.; Riley,~P. Optimization of molecules
  via deep reinforcement learning. \emph{arXiv preprint arXiv:1810.08678}
  \textbf{2018}, \relax
\mciteBstWouldAddEndPunctfalse
\mciteSetBstMidEndSepPunct{\mcitedefaultmidpunct}
{}{\mcitedefaultseppunct}\relax
\EndOfBibitem
\bibitem[Chuai \latin{et~al.}(2018)Chuai, Ma, Yan, Chen, Hong, Xue, Zhou, Zhu,
  Chen, Duan, \latin{et~al.} others]{chuai2018deepcrispr}
Chuai,~G.; Ma,~H.; Yan,~J.; Chen,~M.; Hong,~N.; Xue,~D.; Zhou,~C.; Zhu,~C.;
  Chen,~K.; Duan,~B., \latin{et~al.}  DeepCRISPR: optimized CRISPR guide RNA
  design by deep learning. \emph{Genome biology} \textbf{2018}, \emph{19},
  80\relax
\mciteBstWouldAddEndPuncttrue
\mciteSetBstMidEndSepPunct{\mcitedefaultmidpunct}
{\mcitedefaultendpunct}{\mcitedefaultseppunct}\relax
\EndOfBibitem
\bibitem[Hofacker(2003)]{hofacker2003vienna}
Hofacker,~I.~L. Vienna RNA secondary structure server. \emph{Nucleic acids
  research} \textbf{2003}, \emph{31}, 3429--3431\relax
\mciteBstWouldAddEndPuncttrue
\mciteSetBstMidEndSepPunct{\mcitedefaultmidpunct}
{\mcitedefaultendpunct}{\mcitedefaultseppunct}\relax
\EndOfBibitem
\bibitem[Lorenz \latin{et~al.}(2011)Lorenz, Bernhart, Zu~Siederdissen, Tafer,
  Flamm, Stadler, and Hofacker]{lorenz2011viennarna}
Lorenz,~R.; Bernhart,~S.~H.; Zu~Siederdissen,~C.~H.; Tafer,~H.; Flamm,~C.;
  Stadler,~P.~F.; Hofacker,~I.~L. ViennaRNA Package 2.0. \emph{Algorithms for
  molecular biology} \textbf{2011}, \emph{6}, 26\relax
\mciteBstWouldAddEndPuncttrue
\mciteSetBstMidEndSepPunct{\mcitedefaultmidpunct}
{\mcitedefaultendpunct}{\mcitedefaultseppunct}\relax
\EndOfBibitem
\bibitem[Schaul \latin{et~al.}(2015)Schaul, Quan, Antonoglou, and
  Silver]{schaul2015prioritized}
Schaul,~T.; Quan,~J.; Antonoglou,~I.; Silver,~D. Prioritized experience replay.
  \emph{arXiv preprint arXiv:1511.05952} \textbf{2015}, \relax
\mciteBstWouldAddEndPunctfalse
\mciteSetBstMidEndSepPunct{\mcitedefaultmidpunct}
{}{\mcitedefaultseppunct}\relax
\EndOfBibitem
\bibitem[Yan \latin{et~al.}(2015)Yan, Zhou, and Xue]{yan2015crispr}
Yan,~M.; Zhou,~S.-R.; Xue,~H.-W. CRISPR Primer Designer: Design primers for
  knockout and chromosome imaging CRISPR-Cas system. \emph{Journal of
  integrative plant biology} \textbf{2015}, \emph{57}, 613--617\relax
\mciteBstWouldAddEndPuncttrue
\mciteSetBstMidEndSepPunct{\mcitedefaultmidpunct}
{\mcitedefaultendpunct}{\mcitedefaultseppunct}\relax
\EndOfBibitem
\bibitem[Reuter and Mathews(2010)Reuter, and Mathews]{reuter2010rnastructure}
Reuter,~J.~S.; Mathews,~D.~H. RNAstructure: software for RNA secondary
  structure prediction and analysis. \emph{BMC bioinformatics} \textbf{2010},
  \emph{11}, 1--9\relax
\mciteBstWouldAddEndPuncttrue
\mciteSetBstMidEndSepPunct{\mcitedefaultmidpunct}
{\mcitedefaultendpunct}{\mcitedefaultseppunct}\relax
\EndOfBibitem
\bibitem[Anderson-Lee \latin{et~al.}(2016)Anderson-Lee, Fisker, Kosaraju, Wu,
  Kong, Lee, Lee, Zada, Treuille, and Das]{anderson2016principles}
Anderson-Lee,~J.; Fisker,~E.; Kosaraju,~V.; Wu,~M.; Kong,~J.; Lee,~J.; Lee,~M.;
  Zada,~M.; Treuille,~A.; Das,~R. Principles for predicting RNA secondary
  structure design difficulty. \emph{Journal of molecular biology}
  \textbf{2016}, \emph{428}, 748--757\relax
\mciteBstWouldAddEndPuncttrue
\mciteSetBstMidEndSepPunct{\mcitedefaultmidpunct}
{\mcitedefaultendpunct}{\mcitedefaultseppunct}\relax
\EndOfBibitem
\end{mcitethebibliography}

\end{document}